\documentclass[conference]{IEEEtran}
\IEEEoverridecommandlockouts
\usepackage{cite}
\usepackage{amsmath,amssymb,amsfonts}
\usepackage{algorithmic}
\usepackage{graphicx}
\usepackage{textcomp}
\usepackage{xcolor}
\def\BibTeX{{\rm B\kern-.05em{\sc i\kern-.025em b}\kern-.08em
    T\kern-.1667em\lower.7ex\hbox{E}\kern-.125emX}}
\begin{document}

\title{Enhancing Spatiotemporal Prediction Model using Modular Design and Beyond}

\author{
    Haoyu Pan \textsuperscript{\rm 1} ,
    Hao Wu \textsuperscript{\rm 2},
    Tan Yang\footnote{Corresponding author} \textsuperscript{\rm *,1} \\
    \textsuperscript{\rm 1} School of Computer, Beijing University of Posts and Telecommunications\\
    \textsuperscript{\rm 2}  Institute of Advanced Technology, University of Science and Technology of China\\
    haoyu\_pan@bupt.edu.cn, easylearninghao@gmail.com, tyang@bupt.edu.cn
}



\maketitle

\begin{abstract}
Predictive learning uses a known state to generate a future state over a period of time. It is a challenging task to predict spatiotemporal sequence because the spatiotemporal sequence varies both in time and space. The mainstream method is to model spatial and temporal structures at the same time using RNN-based or transformer-based architecture, and then generates future data by using learned experience in the way of auto-regressive. The method of learning spatial and temporal features simultaneously brings a lot of parameters to the model, which makes the model difficult to be convergent. In this paper, a modular design is proposed, which decomposes spatiotemporal sequence model into two modules: a \textbf{spatial encoder-decoder} and a \textbf{predictor}. These two modules can extract spatial features and predict future data respectively. The spatial encoder-decoder maps the data into a latent embedding space and generates data from the latent space while the predictor forecasts future embedding from past. By applying the design to the current research and performing experiments on KTH-Action and MovingMNIST datasets, we both improve computational performance and obtain state-of-the-art results.
\end{abstract}

\begin{IEEEkeywords}
Spatialtemporal predivitve learning, Representation learning, Deep Learning
\end{IEEEkeywords}

\section{Introduction}
\label{sec:intro}
Space-time dynamical system is common in the real world so related research on spatiotemporal predictive learning has received extensive attention. It is widely used in video prediction\cite{MAU,PhyD}, traffic flow prediction\cite{T-GCN,STG,traffictransformer}, weather forecast\cite{weather1,weather2} and physical simulation system\cite{phy1}. Different from time series prediction, the spatiotemporal sequence at each time step can be regarded as an image (or frame) distributed in higher dimensional feature space. Spatiotemporal predictive learning requires sufficient modeling of time and space structure\cite{survey}.From the perspective of time, the model requires short-term and long-term memory features, while from the perspective of space, it needs to capture both short-range and long-range information.

In current deep neural network research, recurrent neural network (RNN) class models have extraordinary effect on processing time-varying sequence, while convolutional neural network(CNN) are widely used to capture spatial features of images. Therefore, Shi et al. proposed an end-to-end model that combines the architectures of CNN and RNN, called ConvLSTM\cite{convlstm}, for spatio-temporal predictive learning. On the other hand, with the advent of the transformer architecture, an increasing number of sequence models turn to use transformer, like natural language processing(NLP)\cite{transformer,bert}. Transformer has better parallelism than the RNN architecture, as demonstrated in NLP and time series tasks. Then the spatiotemporal analysis models based on transformer\cite{convtransformer,traffictransformer} began to appear. However, there are two deficiencies in both RNN-based and transformer-based architectures:

\textbf{Deficiency I.} During training, the model needs to capture temporal dependence and spatial variance simultaneously. The large number of parameters involved in training and slow calculation will hinder the convergence of the model.

\textbf{Deficiency II.} In terms of spatiotemporal feature encoding and decoding methods, the model with pure CNN architecture performs well in the task of image feature extraction, but performs poorly in the task of image generation. Meanwhile, it is even more difficult for CNN to model time series changes and spatial features at the same time, which leads to distortion of CNN's frame generation results in video sequences. Pixel-level prediction of spatiotemporal sequences is still a challenging task.

We can decouple the spatiotemporal sequence prediction model into three parts: the first part is used to capture the spatial features of the current time step, which we call the spatial encoder. The second part combines the spatial information captured by the first module and passes it in time to predict the spatial features of the future frames, which we call the predictor. The last part decodes the future spatial features generated by the predictor to generate the real future spatiotemporal data, which we call the spatial decoder. Since the spatial encoder and decoder are inverse operations to each other, we can implement it with an encoder-decoder architecture, which we call the spatial encoder-decoder. For example, in both ConvLSTM and ConvTransformer, a pure CNN architecture is used to encode and decode spatial features.

In computer systems, complex systems are often disassembled into relatively simple sub-modules and then combined, and artificial intelligence systems can do the same. In recent studies, including the current outstanding models like PredRNN\cite{predrnn} and 3D-temporal convolutional transformer(TCTN)\cite{TCTN}, both parameters for both spatially dependent and temporally dependent aspects are updated at training time, which leads to inefficiency of the model: a large number of parameters need to be updated each time during training, so that convergence is very slow. To address the deficiency I and II mentioned above, we provide ideas to solve them.

For \textbf{Deficiency I}, we try to decompose a model into two modules: a spatial encoder-decoder module and a predictive learning module, which is called a predictor. The spatial encoder converts data into a latent space. Firstly, we train a static image encoder-decoder, and when we do spatiotemporal sequence prediction, we only need to input spatiotemporal data directly to this spatial encoder-decoder, get the representation tensor of spatiotemporal data at each time step of the sequence, and use this representation tensor instead of the original image. This representation tensor is used to generate the representation tensor of the future data instead of the original image. Finally, the representation tensor generated by the predictor is input to the spatial decoder to obtain the real spatiotemporal data. For \textbf{Deficiency II}, we use Vector Quantised-Variational AutoEncoder(VQ-VAE)\cite{vqvae} to do this task instead of CNN. VQ-VAE is an auto-encoder that has excellent performance in image generation tasks and its architecture fits our needs.

We use two representative spatiotemporal sequence prediction models, PredRNN and TCTN, as the predictor. The former predicts data using an RNN-like architecture and the latter uses a transformer architecture for prediction. The modification of these two different architectures demonstrates the generality of our proposed approach. We do not make major changes to the structure of the original models. The main changes are: on the one hand, the input layer of the original model is modified so that the images are first processed by a pre-trained VQ-VAE encoder before being fed into the model. On the other hand, the parameters of the original model are reduced (the number of layers remained the same, and the hidden units of each layer were halved). Finally, we find experimentally that our approach improves the final results of the model despite the reduction of the predictor parameters.

In summary, the contributions of this paper include the following:
\begin{itemize}
\item Applying a modular design that decomposes sequence prediction model into two parts: spatial decoder-encoder and predictor to making the training of the model easier and more efficient.
\item Utilizing VQ-VAE to encode and generate images, which improves the data generation quality.
\end{itemize}

\section{Related work}
In this section we introduce two different temporal sequence model architectures and analyze their characteristics. On the other hand, a VAE model is introduced.

\textbf{Convolutional RNNs.} Models of ConvRNN-like architectures were first proposed by Shi et al. They use convolutional operations instead of the fully connected layer in FC-LSTM as a way to better capture spatiotemporal correlations in spatiotemporal sequence data. However, due to its simple architecture, ambiguity can occur in the generation of future frames. Many variant models of ConvLSTM have been proposed\cite{trajgru,saconvlstm,mim}, among which PredRNN is an improved ConvLSTM network that is typical of RNN-based approaches. PredRNN adds a spatiotemporal memory unit to the architecture of ConvLSTM and allows the data to be passed along horizontal and vertical directions, enabling the model to better model spatiotemporal correlation. Also RredRNN employs some techniques to make better predictions, such as reverse schedule sampling(RSS), which gradually replaces the real data with predicted data as input in a multi-step iterative prediction process. The shortcomings of ConvRNNs are that they are limited by the RNN architecture, have average prediction results when the sequences are long, and are computationally intensive.

\textbf{Convolutional Tranformers.} Transformer research has gained significant results in both natural language processing and computer vision fields, transformer-based models in spatiotemporal sequence prediction tasks have gradually become more numerous. The standard transformer uses stacked transformer blocks with spatial attention and temporal attention for spatiotemporal sequence prediction, however, the drawback is the same as FC-LSTM, limited by the fully connected layers used, the model cannot capture short-term denpency well. ConvTransformer\cite{convtransformer} applies the transformer structure to the domain of spatiotemporal sequential data, using multi-headed convolutional attention instead of the traditional one-dimensional attention mechanism to capture spatiotemporal features. convTransformer performs better in predicting intermediate content with known context, but performs mediocrely for future information. Inspired by the multi-headed convolutional attention in ConvTransformer, Yang et al. proposed TCTN\cite{TCTN}, a 3D convolutional transformer-based spatiotemporal sequence prediction model.TCTN uses a transformer-based encoder with temporal convolutional layers to be used to capture both short-term and long-term dependencies, achieving results comparable to PredRNN in some metrics. The results are comparable to those of PredRNN in some metrics.

\textbf{VQ-VAE.} Vector quantised variational autoencoder(VQ-VAE), as known as a neural discrete representation learning method, is a very effective auto encoder (AE) model\cite{vqvae}. It consists of an encoder and a decoder. In the image generation task, the image is input to the encoder and a latent embedding vector $z$ is output. VQ-VAE also maintains a dictionary, which is a vector of length $D$. The nearest vector $z'$ is obtained by performing a nearest neighbor search on $z$ in the dictionary, and $z'$ is used as the final encoding result instead of $z$. VQ-VAE uses a CNN with residual units to perform a reduction of the encoding result $z'$, to fit the distribution of the encoding. When training VQ-VAE, since the process of discrete encoding has no gradient, VQ-VAE uses Straight-Through Estimator instead of the usual loss function. The model has great power to construct images, videos and speech.

\section{Methods}
\subsection{Overview}
\begin{figure*}[t]
  \centering
  \includegraphics[width=\textwidth]{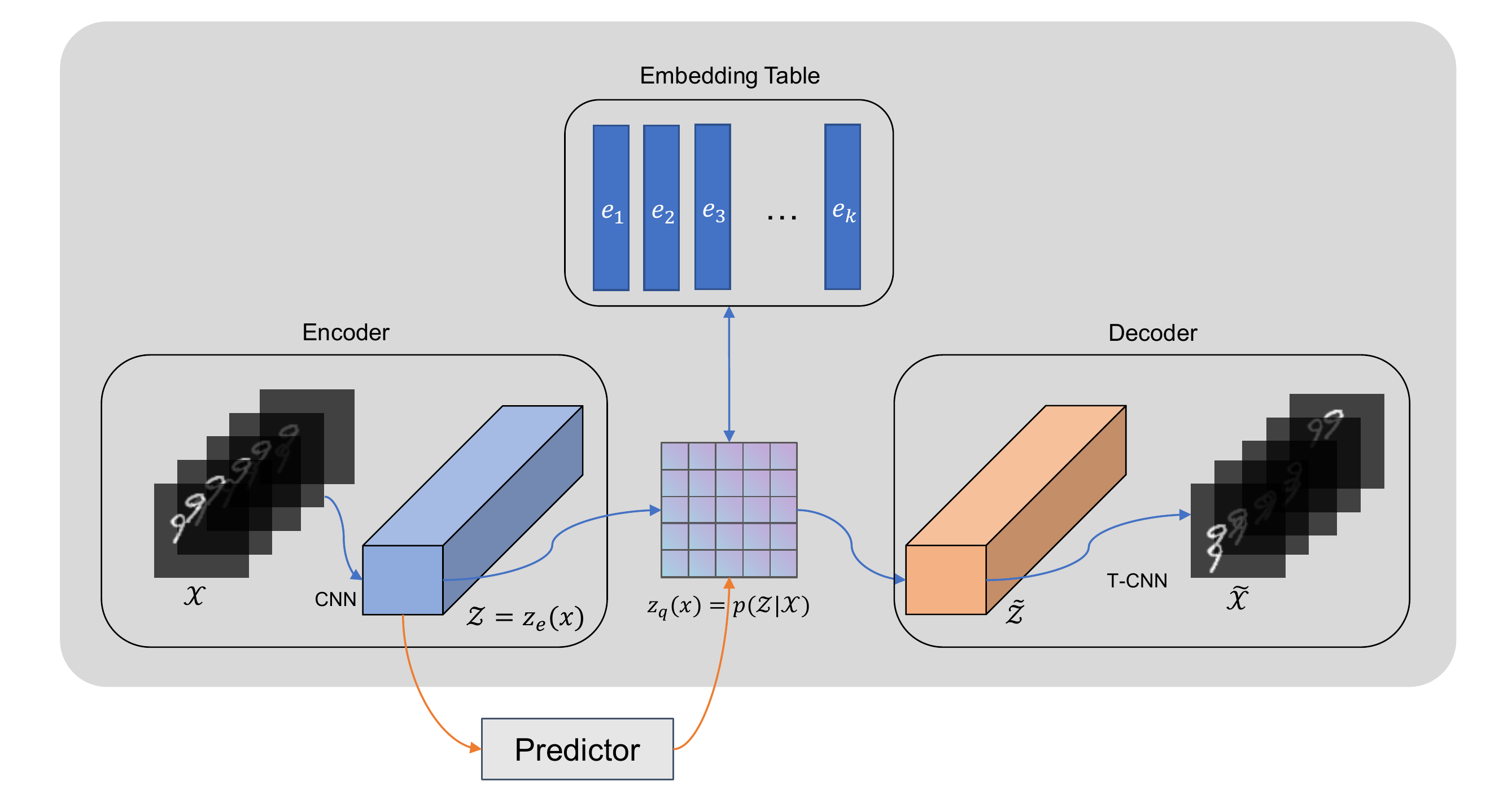}
  \vskip -0.1in
  \caption{Pipeline of the modularly designed spatiotemporal sequence prediction model. The blue line indicates the data flow of the spatial encoder-decoder module, and the orange line indicates the data flow of the predictor module.}
  
  \label{fig:pipeline}
\end{figure*}

A spatiotemporal sequence of data can be represented as a length-$S$ sequence $\mathcal{X} = \{\mathcal{X}_1, \mathcal{X}_2, ..., \mathcal{X}_S\}$, each $\mathcal{X}_t \in \mathbb{R}^{W\times H \times D}, t = 1, 2, ..., S$, means $\mathcal{X}_t$ has dimensions of $H\times W$ and depth of $D$. The spatiotemporal model is to use $\mathcal{X}$ to predict length-m sequence ${\mathcal{\tilde{X}}} = \{\mathcal{\tilde{X}}_1, \mathcal{\tilde{X}}_2, ..., \mathcal{\tilde{X}}_T\} $ which is conitunous after $\mathcal{X}$. This means that the spatiotemporal predictive learning can be expressed as an optimization problem with maximum conditional probability:
\begin{equation}
{\mathcal{\tilde{X}}} = \mathop{\arg\min}\limits_{\mathcal{X}_1 ..., \mathcal{X}_T} p(\mathcal{X}_1, ..., \mathcal{X}_T\ | \mathcal{X})
\end{equation}
The structure of our proposed method is shown in Figure 1. In this paper, instead of using the original sequence data as the input to the predictor, we use the embedding $\mathcal{Z} = \{\mathcal{Z}_1, \mathcal{Z}_2, ..., \mathcal{Z}_S\}, \text{where}\, \mathcal{Z}_i = \text{Encoder}(\mathcal{X}_i), i = 1, 2, ..., S$.

In CNN, the size of the original image affects the number of convolutional kernel slides, which in turn affects the computation speed. The new embedding $\mathcal{Z}$ of the original image, is smaller than  $\mathcal{X}$ in size after encoder dimensionality reduction, and the number of convolution kernel computations is reduced, thus improving the computational speed of the predictor. On the other hand, $\mathcal{Z}$ contains richer features because encoder superimposes the extracted features to the channel dimension of $\mathcal{Z}$.

Thus, we turn the original task of predicting the embedding $\mathcal{\tilde{X}}$ by $\mathcal{X}$ into predicting the embedding $\mathcal{\tilde{Z}}$ from the embedding $\mathcal{X}$:

\begin{equation}
{\mathcal{\tilde{Z}}} = \mathop{\arg\min}\limits_{\mathcal{Z}_1 ..., \mathcal{Z}_T} p(\mathcal{Z}_1, ..., \mathcal{Z}_T\ | \mathcal{Z})
\end{equation}

\subsection{Spatial Encoder-Decoder}
Spatial Encoder-Decoder encodes the data $\mathcal X_i$ to $\mathcal{Z}_i$ at each time step in $\mathcal X$ by Encoder:
\begin{equation}
\mathcal{Z}_i = \rm{Encoder}(\mathcal{X}_i), \mathcal{Z}_i \in \mathbb{R}^{H'\times W' \times D'}
\end{equation}
Then $\mathcal{Z}_i$ is transformed into a one-hot embedding through the posterior distribution probability distribution.
\begin{equation}
p(z=k|x)=
\begin{cases}
&1, \text{for}\, k = \arg\min_j{ \|z_e(x) - e_j\|_2} \\
&0, \text{otherwise} 
\label{eq_assign}
\end{cases}
\end{equation}
The nearest neighbor algorithm is used to obtain the potential encoding vector of the corresponding index in the embedding table:
\begin{equation}
z_q(x) = e_k, \text{where} \, k = \arg \min_j \Vert z_e(x) - e_j \Vert_2
\end{equation}
Finally, the decoder is then used to reduce $z_q$ to a real-world image: 
\begin{equation}
\mathcal{\tilde{X}}_i = \text{Decoder}(\mathcal{\tilde{Z}}_i) 
\end{equation}

\subsection{Predictor}
As mentioned in Section \ref{sec:intro}, we use two different architectures, respectively transformer-based and RNN-based models, to validate the approach proposed in this paper. We have selected typical representatives of these two models, which are PredRNN and TCTN, and they both achieve SOTA performance in most metrics.

 \begin{figure}[htbp]
  \centering
  \includegraphics[width=\columnwidth]{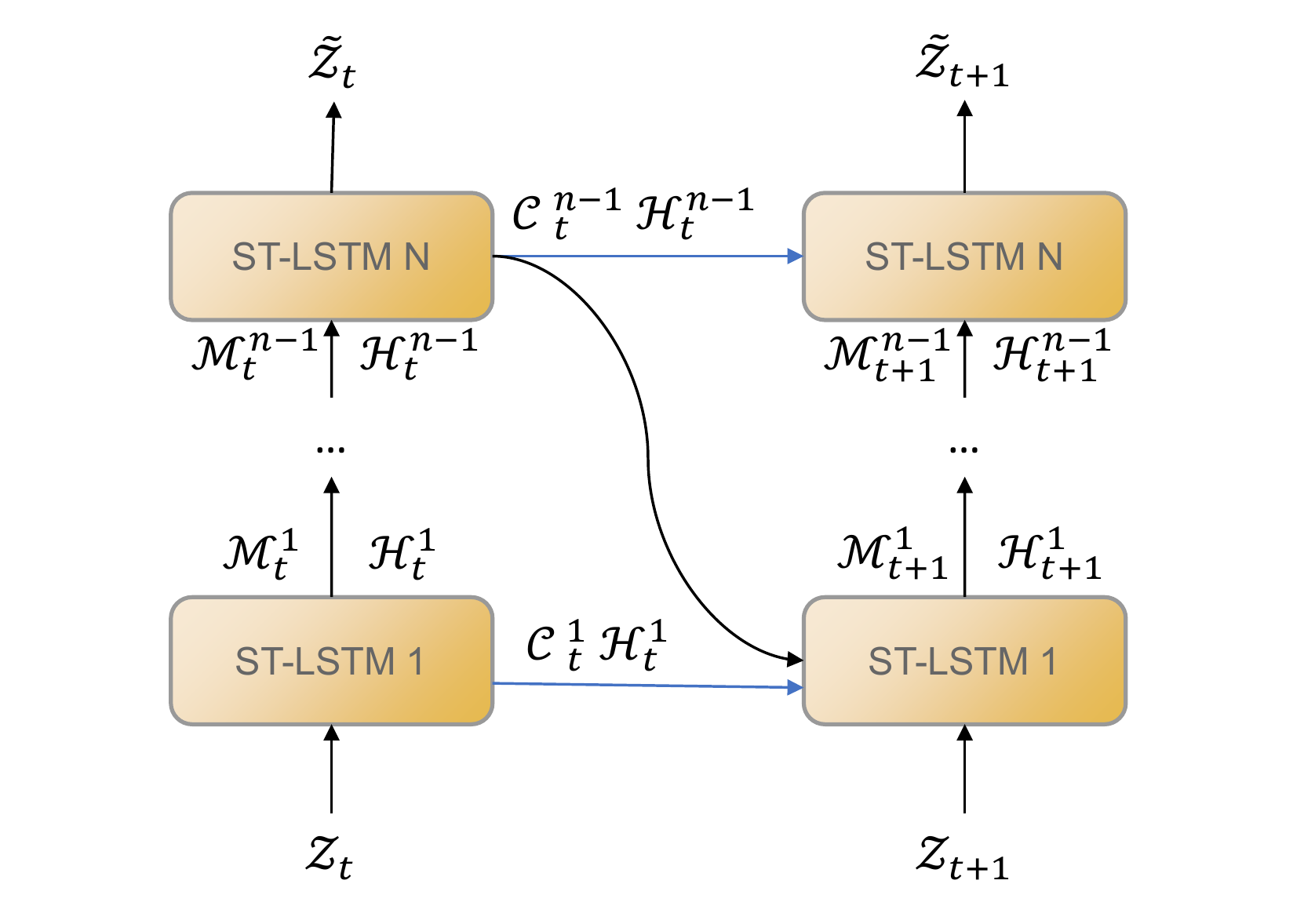}
  \caption{Architecture of PredRNN. The temporal memory $\mathcal{M}$ is passed through the layers along the black arrows with time $t$ Temporal memory $\mathcal{C}$ is passed horizontally in the same layer as time changes. The hidden state $\mathcal{H}$ is passed along the zigzag direction in the model.}
  \label{fig:predrnn}
\end{figure}

\textbf{PredRNN.} The model structure of PredRNN is shown in Figure \ref{fig:predrnn}. PredRNN is a model of recurrent neural network architecture, where each recurrent layer consists of multiple spatiotemporal LSTM units stacked. In PredRNN, a new state variable $\mathcal{M}$ is defined in addition to the memory state $\mathcal{C}$ and the hidden state $\mathcal{H}$ in the standard LSTM network. $\mathcal{C}$ is a temporal memory unit that propagates with time $t$; The hidden state $\mathcal{H}$ can be simultaneously passed along the horizontal direction and passed up along the vertical direction. Meanwhile, in addition to the input, forgetting and output gates in the standard LSTM, input modulation gate $g'$, input gate $i'$ and forgetting gate $f'$ are introduced to update the spatiotemporal memory unit $\mathcal{M}$, which is calculated as following:
\begin{small} 
\begin{equation}
  \begin{split}
  g_t & = \tanh(W_{xg} \ast \mathcal{X}_t + W_{hg} \ast \mathcal{H}_{t-1}^l) \\
  i_t & = \sigma(W_{xi} \ast \mathcal{X}_t + W_{hi} \ast \mathcal{H}_{t-1}^l) \\
  f_t & = \sigma(W_{xf} \ast \mathcal{X}_t + W_{hf} \ast \mathcal{H}_{t-1}^l) \\
  \mathcal{C}_t^l & = f_t \odot \mathcal{C}_{t-1}^l + i_t \odot g_t \\
  g_t^\prime & = \tanh(W_{xg}^\prime \ast \mathcal{X}_t + W_{mg} \ast \mathcal{M}_t^{l-1}) \\
  i_t^\prime & = \sigma(W_{xi}^\prime \ast \mathcal{X}_t + W_{mi} \ast \mathcal{M}_t^{l-1}) \\
  f_t^\prime & = \sigma(W_{xf}^\prime \ast \mathcal{X}_t + W_{mf} \ast \mathcal{M}_t^{l-1}) \\
  \mathcal{M}_t^l & = f_t^\prime \odot \mathcal{M}_t^{l-1} + i_t^\prime \odot g_t^\prime \\
  o_t & = \sigma(W_{xo} \ast \mathcal{X}_t + W_{ho} \ast \mathcal{H}_{t-1}^l + W_{co} \ast \mathcal{C}_t^l + W_{mo} \ast \mathcal{M}_t^l) \\ 
  \mathcal{H}_t^l & = o_t \odot \tanh(W_{1\times1} \ast [\mathcal{C}_t^l, \mathcal{M}_t^l]).\\
  \end{split} 
\end{equation}
\end{small} 

\textbf{TCTN.}The flow chart of TCTN calculation is shown in Figure \ref{fig:tctn}. TCTN is a transformer-based spatiotemporal sequence prediction model that uses 3D convolutional operations instead of linear operations in the standard transformer structure to compute correlations between spatiotemporal data. Since the transformer is insensitive to the position of the sequence data, the position information is represented by adding the position embedding $P\in \mathbb{R}^{H'\times W' \times C'}$ to the input data, and $P$ is calculated as follows:

\begin{figure}[htbp]
\centering
\includegraphics[width=\columnwidth]{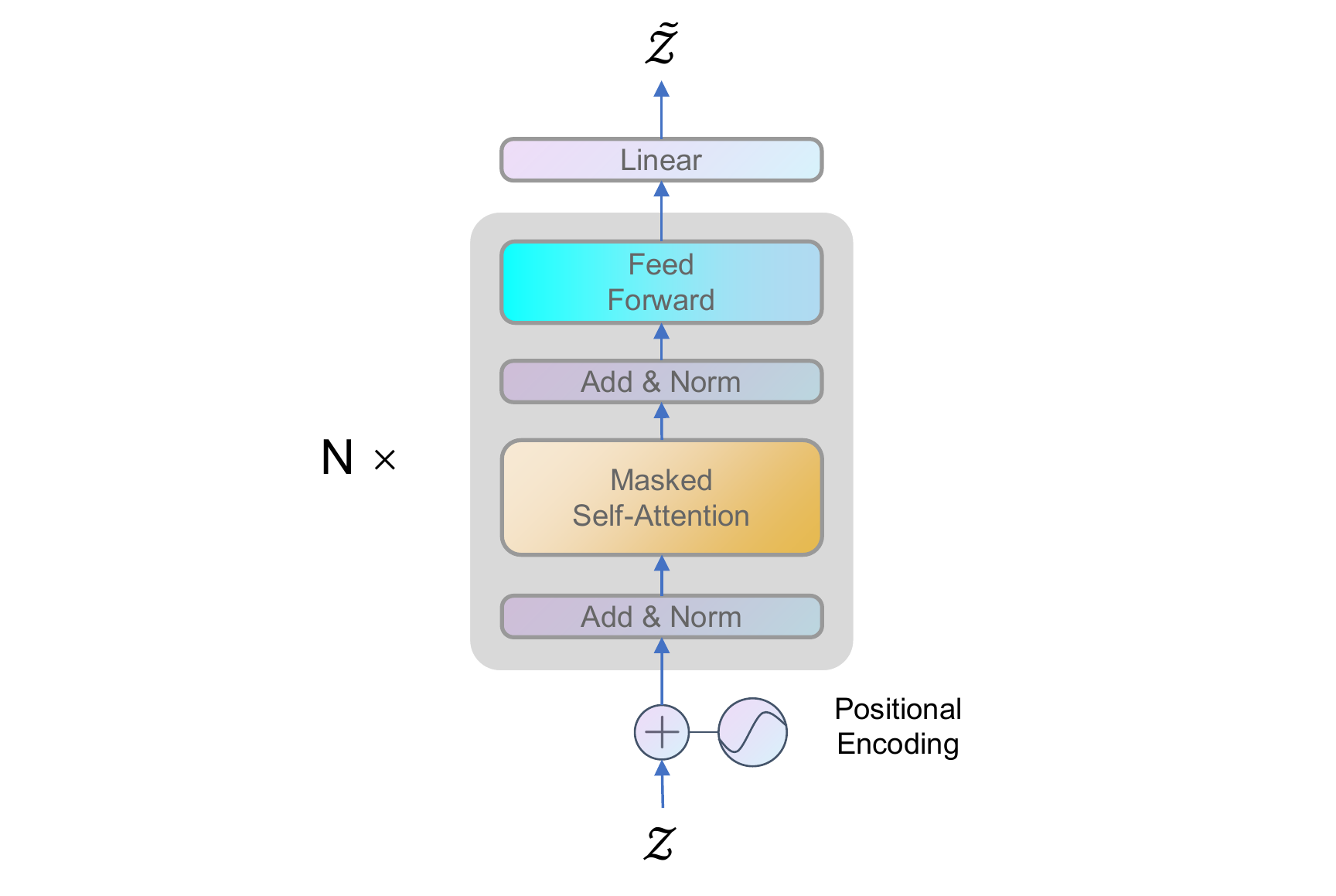}
\caption{Overview of TCTN Model Architecture. The model can be considered as a transformer decoder using 3D convolutional operations instead of linear operations.}
\label{fig:tctn}
\vspace{-10pt}
\end{figure}

\begin{equation}
\begin{split}
   P_{j,h,w,2d} & = \sin(j/10000^{2d/D}), \\
   P_{j,h,w,2d+1} & = \cos(j/10000^{2d/D}), \\
   E & = \mathcal{Z} + P
\end{split}
\end{equation}
After getting the input $E$ with position embedding, it is fed to Spatiotemporal Decoder to perform the prediction. Specifically, after first layer normalizing E, the computation of the matrices $Q\in \mathbb{R}^{S\times H' \times W' \times D'}$, $K\in \mathbb{R}^{S \times H' \times W' \times D'}$ and $V\in \mathbb{R}^{T\times H' \times W' \times D'}$, and compute the scaled dot-product attention matrix $A$.
\begin{equation}
\begin{split} 
\hat{E} &= \text{LN}(E), \\
Q &= W_q * \hat{E}, \\
K &= W_k * \hat{E}, \\
V &= W_v * \hat{E}, \\
A &= \text{Softmax}\bigg(\text{Mask}(\frac{QK^T}{\sqrt{D'}})\bigg)V
\end{split} 
\end{equation}
The difference from the standard attention mechanism is that the parameter matrices $W_q$, $W_k$, $W_w$ of TCTN are 3D convolutional kernels instead of the weights of the linear layers. Then, $A$ is added to the original input E by residual concatenation and layer normalized to obtain $\hat{S}$, which is used as the input of a feed forward network(FFN) layer. The feed forward network consists of two 3D convolutional layers and a leaky relu activation function, and the final output predicts the encoding of the spatiotemporal sequence $\mathcal{Z}$:

\begin{equation}
\begin{split}
\hat{A} &= \text{LN}(A) \\
S &= E + \hat{A} \\
\hat{S} &= \text{LN}(S) \\
\tilde{\mathcal{Z}} &= W_1 * (\text{LReLU}(W_2 * \hat{S}))
\end{split} 
\end{equation}
$\text{LReLU}(\cdot)$ denotes the leaky relu activation function and both $W_1$ and $W_2$ are 3D convolution kernels.

\section{Experiments}
\begin{figure*}[t]
  \centering
  \includegraphics[width=\textwidth]{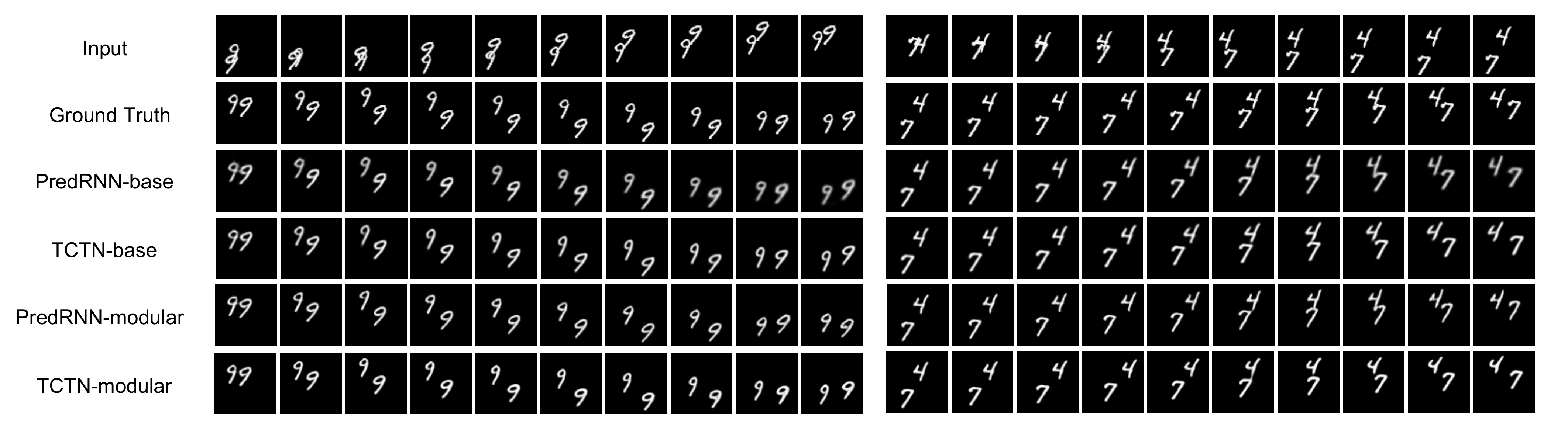}
  \caption{Performance comparison of different models on MovingMNIST dataset.}
  \label{fig:mnist}
\end{figure*}
\subsection{Dataset}
For fair comparison, we use the dataset used in both TCTN and PredRNN, they are:
\begin{itemize}
  \item MovingMNIST: MovingMNIST is to randomly select one or more static handwritten digital images from the MNIST handwritten digital dataset, and give these numbers an initial change speed and angle, so that the numbers keep this speed and angle in a certain range of motion, so as to generate a space-time sequence. In this article, we are consistent with the data set used by PredRNN\cite{predrnn} and TCTN\cite{TCTN}: the generated sequence of 10,000 is 20 frames, and each frame is a 64$\times$ 64 gray-scale image, with the first 10 frames as input and the last 10 frames as prediction targets.We then use a sequence of 5,000 lengths as the test set.
  \item KTH-Action\cite{kth}: KTH-Action is a real-world dataset. It captures video sequences of six different categories of actions performed by 25 people in four different scenarios. The average length of the video sequences was 4 seconds, with a frame rate of 25FPS and a resolution of 160 $\times$ 120. We follow the same settings with \cite{kth_set}: utilizing people numbered 1-16 as the training set, and the rest as the test set. Each frame is rescaled into a 128 $\times$ 128 grayscale image. A sample is a sequence of images containing 20 frames, with the first 10 frames as input and the last 10 frames as target.
  
\end{itemize}

\subsection{Compared Metrics}
We divide the experiments into two aspects: prediction quality comparison and computational performance comparison.

To compare the generated data results, we keep the parameters of the original model and our modified model the same, only the input data are changed. The input data of the basic reference model are real images, while our modified model, called PredRNN-modular or TCTN-modular, uses the latent embedding $\mathcal{Z}$ generated by the encoder of VQ-VAE as input. We use the following metrics to measure the quality:

\begin{itemize}
  \item Learned Perceptual Image Patch Similarity(LPIPS). This metric is used to measure the difference between two images by learning the reverse mapping of the generated image to Ground Truth, and forcing the generator to learn the reverse mapping of reconstructing the real image from the fake image and prioritizing the perceptual similarity between them\cite{lpips}. LPIPS is more consistent with human perception than traditional methods. A lower value of LPIPS indicates that the two images are more similar.
  \item Structural Similarity Index Measure (SSIM). SSIM measures image similarity in terms of brightness, contrast and structure respectively. The bigger the SSIM, the better the result.
\end{itemize}

In order to compare the computational performance of different models, we fine-tune the original model by increasing the number of parameters, using the LPIPS metrics in the MNIST dataset as a benchmark until the final prediction is essentially the same as our model. Then we calculate the training consumption (time required for training per 100 iterations, GPU memory usage) and the total number of parameters for the model after tuning the parameters.

\subsection{Implementation Detail}
All basic models and predictors of modular models keep the same setup with TCTN(6 decoder layers, self-attention head, 64 hidden units each layers) and PredRNN (4 layers, each layer contains 64 hidden units) as controls, and train each model for 70,000 iterations to choose the best.


\subsubsection{MovingMNIST} 

\begin{figure}[!h]
  \centering
  \includegraphics[width=\columnwidth]{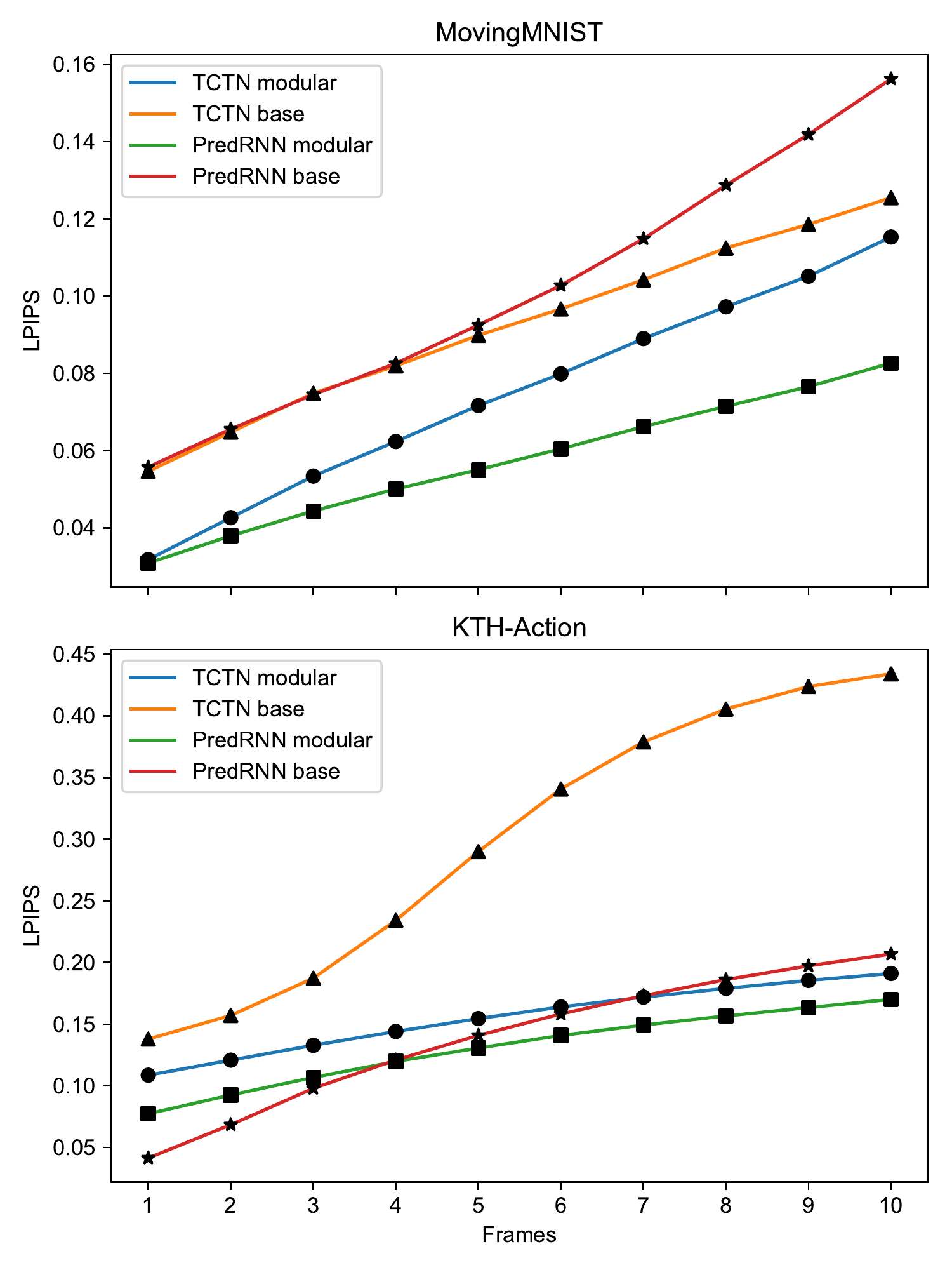}
  \caption{Frame-wise LPIPS results on test set. \textbf{Above:} MovingMNIST. Both models using the method proposed in this paper perform better. \textbf{Below:} KTH-Action. Our method is effective on TCTN and comparable to the base model on PredRNN. However, the result generated by our method has a smoother variance.}
  \label{fig:mnistlpips}
  \vspace{-10pt}
\end{figure}

\begin{table}[htbp]
\caption{Quality of predictions on MovingMNIST dataset.}
	\centering
	\begin{tabular}{lcccc}
	    \hline
		\multicolumn{1}{c}{MODEL} & \multicolumn{1}{c}{LPIPS(↓)} & \multicolumn{1}{c}{SSIM(↑)}   \\
        \hline
		PredRNN-base & 0.102   & 0.871   \\
		\textbf{PredRNN-modular} & \textbf{0.055}  & \textbf{0.884}  \\
		\hline
		TCTN-base & 0.094 & \textbf{0.861}  \\
		\textbf{TCTN-modular} & \textbf{0.077} & 0.859    \\
		\hline
	\end{tabular}%
    \label{table:mnist-performance}%
\end{table}


The experiment results of the MovingMNIST dataset are shown in Table \ref{table:mnist-performance} and Figure \ref{fig:mnist}. In the MovingMNIST dataset, our improvements to both models achieve significant improvements in the LPIPS metric. This indicates that the spatiotemporal data generated using our approach has better human readability. In terms of SSIM, the results are comparable to the original model.

    

\begin{table}[htbp]
	\caption{Quality of predictions on KTH-Action dataset.}
	\centering
	\begin{tabular}{lcccc}
        \hline
		\multicolumn{1}{c}{MODEL} &  \multicolumn{1}{c}{LPIPS(↓)} &  \multicolumn{1}{c}{SSIM(↑)} \\ 
		\hline
		PredRNN-base & \textbf{0.124} & \textbf{0.854}  \\
		\textbf{PredRNN-modular} & 0.129  & 0.848 \\
		\hline
		TCTN-base & 0.298 & 0.645 \\
		\textbf{TCTN-modular} & \textbf{0.155}  & \textbf{0.832} \\
		\hline
	\end{tabular}%
    \label{table:kth-performance}%
\end{table}


\begin{figure*}[htpb]
  \centering
  \includegraphics[width=\textwidth]{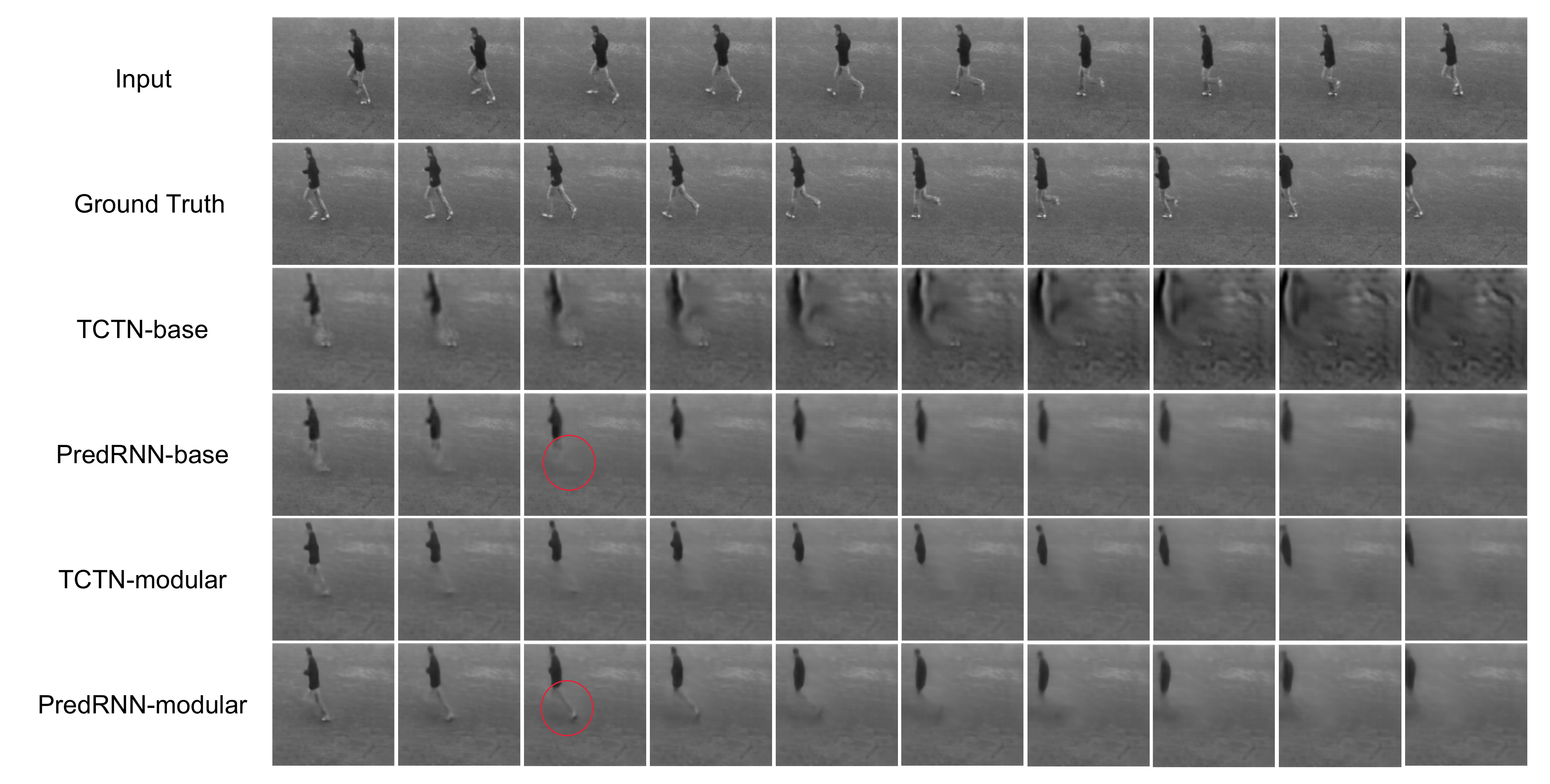}
  \caption{Performance comparison of different models on KTH-Action dataset.}
  \label{fig:kth}
\end{figure*}

\subsubsection{KTH-Action} 

The experiment result for the KTH-Action dataset are shown in the Figure \ref{fig:kth} and Table \ref{table:kth-performance}. Differs from the MovingMNIST dataset in that the spatiotemporal dynamics of the KTH-Action dataset vary relatively continuously and has more noise and stochastic variation because it is a real-world picture. Therefore, the modular approach propose in this paper has different effects on the two model enhancements.

For PredRNN, the modular design do not significantly improve the basic model, and all metrics for comparison are similar to the original model.

For TCTN, the base model cannot capture the relatively complex spatiotemporal variations in KTH-Action well because the model is too simple.
It should be noted that in the original paper, the image size of the KTH-Action used is $64\times 64$\cite{TCTN}. The modular design significantly enhances the effectiveness of TCTN in small number of parameters and large sample size scenarios.

\subsubsection{Computational Performance} 
\begin{table*}
	\caption{Computational performance comparison. }
	\centering
	\begin{tabular}{lcccc}
        \hline
		\multicolumn{1}{c}{MODEL} & \multicolumn{1}{c}{LPIPS} & \multicolumn{1}{c}{Params(M)} & \multicolumn{1}{c}{Mem(GB)} & \multicolumn{1}{c}{Training Cost(s)} \\
        \hline
		PredRNN-large & 0.076 & 27.79 & 5.231 & 91.8  \\
		\textbf{PredRNN-modular} & 0.057 & 6.60 + 2.17 & 2.197 & \textbf{70.3} \\
		\hline
		TCTN-large & 0.078 & 25.52 & 2.719  & 93.0  \\
		\textbf{TCTN-modular} & 0.077 & 7.00 + 2.17 & 2.234 & \textbf{39.5} \\
		\hline
	\end{tabular}%
    \label{table:computational-performance}%
\end{table*}

We follow the setup in the previous section to compare the computational performance of the models by testing the cost of the models when they are trained on the MNIST dataset. By increasing the models until the original model and our model achieve the same results on LPIPS. Finally, the number of layers of PredRNN and TCTN remain the same and the hidden units per layer are doubled, and the results are shown in the table \ref{table:computational-performance}. 

Mem contains the memory occupied by both the training data and the model itself. Training cost indicates the time consumed per 100 iterations. All experiments are conducted on a single NVIDIA Tesla V100 GPU with 32G memory. In our model, the number of parameters denotes the sum of the parameters of the pre-trained VQ-VAE model and the predictor parameters, e.g., 6.60 + 2.17 indicates that the number of predictor parameters is 6.60M and the number of VQ-VAE parameters is 2.17M. The parameters of VQ-VAE are no longer updated during training, so there is no need to calculate the gradient.

In terms of the number of parameters, the size of the modular models are 31.6\%(PredRNN) and 35.9\%(TCTN) of the original models, respectively. This results in a 58.0\% reduction in memory usage and a 23.4\% improvement in training speed for PredRNN during training. For TCTN, there are a 17.8\% reduction in memory usage and a 57.5\% increase in training speed.

\section{Conclusion}
In this paper, we propose a general method to enhance the performance of existing spatiotemporal sequence prediction models: decomposing the model into two separate parts, which are called spatial encoder-decoder and predictor. We use a pre-trained VQ-VAE model as the spatial encoder-decoder. The output of the encoder at VQ-VAE is used as the input of the predictor. Finally, real world images are generated by the decoder of VQ-VAE.

In general, we obtain competitive results with the modular design. Some of the metrics are improved significantly on different datasets. Typically, LPIPS indicates that our method has better human readability in generating prediction images and improves the prediction performance. The result on the KTH-Action dataset shows that using a modular design is effective in improving the performance of models that perform poorly on datasets with large sample sizes.

There are many possible attempts at the spatial encoder-decoder, such as VAE, or representation of ViT\cite{ViT}, etc. Meanwhile, more and more researches are turning to big models\cite{videogpt} with massive parameters, the basic idea of this paper is quite opposite to big models. We hope that the ideas in this paper will lead to some new thinking about big models.

\bibliographystyle{IEEEtran}
\bibliography{refs}

\begin{thebibliography}{10}
\providecommand{\url}[1]{#1}
\csname url@samestyle\endcsname
\providecommand{\newblock}{\relax}
\providecommand{\bibinfo}[2]{#2}
\providecommand{\BIBentrySTDinterwordspacing}{\spaceskip=0pt\relax}
\providecommand{\BIBentryALTinterwordstretchfactor}{4}
\providecommand{\BIBentryALTinterwordspacing}{\spaceskip=\fontdimen2\font plus
\BIBentryALTinterwordstretchfactor\fontdimen3\font minus
  \fontdimen4\font\relax}
\providecommand{\BIBforeignlanguage}[2]{{%
\expandafter\ifx\csname l@#1\endcsname\relax
\typeout{** WARNING: IEEEtran.bst: No hyphenation pattern has been}%
\typeout{** loaded for the language `#1'. Using the pattern for}%
\typeout{** the default language instead.}%
\else
\language=\csname l@#1\endcsname
\fi
#2}}
\providecommand{\BIBdecl}{\relax}
\BIBdecl

\bibitem{MAU}
Z.~Chang, X.~Zhang, S.~Wang, S.~Ma, Y.~Ye, X.~Xinguang, and W.~Gao, ``Mau: A
  motion-aware unit for video prediction and beyond,'' in \emph{Advances in
  Neural Information Processing Systems}, M.~Ranzato, A.~Beygelzimer,
  Y.~Dauphin, P.~Liang, and J.~W. Vaughan, Eds., vol.~34.\hskip 1em plus 0.5em
  minus 0.4em\relax Curran Associates, Inc., 2021, pp. 26\,950--26\,962.

\bibitem{PhyD}
V.~{Le Guen} and N.~{Thome}, ``{Disentangling Physical Dynamics from Unknown
  Factors for Unsupervised Video Prediction},'' \emph{arXiv e-prints}, p.
  arXiv:2003.01460, Mar. 2020.

\bibitem{T-GCN}
L.~{Zhao}, Y.~{Song}, C.~{Zhang}, Y.~{Liu}, P.~{Wang}, T.~{Lin}, M.~{Deng}, and
  H.~{Li}, ``{T-GCN: A Temporal Graph ConvolutionalNetwork for Traffic
  Prediction},'' \emph{arXiv e-prints}, p. arXiv:1811.05320, Nov. 2018.

\bibitem{STG}
Q.~Zhang, J.~Chang, G.~Meng, S.~Xiang, and C.~Pan, ``Spatio-temporal graph
  structure learning for traffic forecasting,'' \emph{Proceedings of the AAAI
  Conference on Artificial Intelligence}, vol.~34, pp. 1177--1185, 04 2020.

\bibitem{traffictransformer}
L.~Cai, K.~Janowicz, G.~Mai, B.~Yan, and R.~Zhu, ``Traffic transformer:
  Capturing the continuity and periodicity of time series for traffic
  forecasting,'' \emph{Transactions in GIS}, vol.~24, 06 2020.

\bibitem{weather1}
A.~McGovern, D.~J. Gagne, J.~K. Williams, R.~A. Brown, and J.~B. Basara,
  ``Enhancing understanding and improving prediction of severe weather through
  spatiotemporal relational learning,'' \emph{Machine Learning}, vol.~95,
  no.~1, pp. 27--50, Apr 2014.

\bibitem{weather2}
M.~Jamaly and J.~Kleissl, ``Spatiotemporal interpolation and forecast of
  irradiance data using kriging,'' \emph{Solar Energy}, vol. 158, pp. 407--423,
  2017.

\bibitem{phy1}
V.~Aryai and M.~Mahmoodian, ``Spatial-temporal reliability analysis of
  corroding cast iron water pipes,'' \emph{Engineering Failure Analysis},
  vol.~82, pp. 179--189, 2017.

\bibitem{survey}
X.~{Shi} and D.-Y. {Yeung}, ``{Machine Learning for Spatiotemporal Sequence
  Forecasting: A Survey},'' \emph{arXiv e-prints}, p. arXiv:1808.06865, Aug.
  2018.

\bibitem{convlstm}
X.~SHI, Z.~Chen, H.~Wang, D.-Y. Yeung, W.-k. Wong, and W.-c. WOO,
  ``Convolutional lstm network: A machine learning approach for precipitation
  nowcasting,'' in \emph{Advances in Neural Information Processing Systems},
  C.~Cortes, N.~Lawrence, D.~Lee, M.~Sugiyama, and R.~Garnett, Eds.,
  vol.~28.\hskip 1em plus 0.5em minus 0.4em\relax Curran Associates, Inc.,
  2015.

\bibitem{transformer}
A.~Vaswani, N.~Shazeer, N.~Parmar, J.~Uszkoreit, L.~Jones, A.~N. Gomez,
  L.~Kaiser, and I.~Polosukhin, ``Attention is all you need,'' in
  \emph{Proceedings of the 31st International Conference on Neural Information
  Processing Systems}, ser. NIPS'17.\hskip 1em plus 0.5em minus 0.4em\relax Red
  Hook, NY, USA: Curran Associates Inc., 2017, p. 6000–6010.

\bibitem{bert}
J.~{Devlin}, M.-W. {Chang}, K.~{Lee}, and K.~{Toutanova}, ``{BERT: Pre-training
  of Deep Bidirectional Transformers for Language Understanding},'' \emph{arXiv
  e-prints}, p. arXiv:1810.04805, Oct. 2018.

\bibitem{convtransformer}
Z.~{Liu}, S.~{Luo}, W.~{Li}, J.~{Lu}, Y.~{Wu}, S.~{Sun}, C.~{Li}, and
  L.~{Yang}, ``{ConvTransformer: A Convolutional Transformer Network for Video
  Frame Synthesis},'' \emph{arXiv e-prints}, p. arXiv:2011.10185, Nov. 2020.

\bibitem{predrnn}
Y.~Wang, H.~Wu, J.~Zhang, Z.~Gao, J.~Wang, P.~Yu, and M.~Long, ``Predrnn: A
  recurrent neural network for spatiotemporal predictive learning,'' \emph{IEEE
  Transactions on Pattern Analysis and Machine Intelligence}, vol.~PP, pp.
  1--1, 04 2022.

\bibitem{TCTN}
Z.~{Yang}, X.~{Yang}, and Q.~{Lin}, ``{TCTN: A 3D-Temporal Convolutional
  Transformer Network for Spatiotemporal Predictive Learning},'' \emph{arXiv
  e-prints}, p. arXiv:2112.01085, Dec. 2021.

\bibitem{vqvae}
A.~van~den Oord, O.~Vinyals, and k.~kavukcuoglu, ``Neural discrete
  representation learning,'' in \emph{Advances in Neural Information Processing
  Systems}, I.~Guyon, U.~V. Luxburg, S.~Bengio, H.~Wallach, R.~Fergus,
  S.~Vishwanathan, and R.~Garnett, Eds., vol.~30.\hskip 1em plus 0.5em minus
  0.4em\relax Curran Associates, Inc., 2017.

\bibitem{trajgru}
X.~Shi, Z.~Gao, L.~Lausen, H.~Wang, D.-Y. Yeung, W.-k. Wong, and W.-c. WOO,
  ``Deep learning for precipitation nowcasting: A benchmark and a new model,''
  in \emph{Advances in Neural Information Processing Systems}, I.~Guyon, U.~V.
  Luxburg, S.~Bengio, H.~Wallach, R.~Fergus, S.~Vishwanathan, and R.~Garnett,
  Eds., vol.~30.\hskip 1em plus 0.5em minus 0.4em\relax Curran Associates,
  Inc., 2017.

\bibitem{saconvlstm}
Z.~Lin, M.~Li, Z.~Zheng, Y.~Cheng, and C.~Yuan, ``Self-attention convlstm for
  spatiotemporal prediction,'' \emph{Proceedings of the AAAI Conference on
  Artificial Intelligence}, vol.~34, no.~07, pp. 11\,531--11\,538, Apr. 2020.

\bibitem{mim}
Y.~Wang, J.~Zhang, H.~Zhu, M.~Long, J.~Wang, and P.~S. Yu, ``Memory in memory:
  A predictive neural network for learning higher-order non-stationarity from
  spatiotemporal dynamics,'' in \emph{2019 IEEE/CVF Conference on Computer
  Vision and Pattern Recognition (CVPR)}, 2019, pp. 9146--9154.

\bibitem{kth}
C.~Schuldt, I.~Laptev, and B.~Caputo, ``Recognizing human actions: a local svm
  approach,'' in \emph{Proceedings of the 17th International Conference on
  Pattern Recognition, 2004. ICPR 2004.}, vol.~3, 2004, pp. 32--36 Vol.3.

\bibitem{kth_set}
R.~Villegas, J.~Yang, S.~Hong, X.~Lin, and H.~Lee, ``Decomposing motion and
  content for natural video sequence prediction,'' in \emph{ICLR}, 2017.

\bibitem{lpips}
R.~Zhang, P.~Isola, A.~A. Efros, E.~Shechtman, and O.~Wang, ``The unreasonable
  effectiveness of deep features as a perceptual metric,'' in \emph{2018
  IEEE/CVF Conference on Computer Vision and Pattern Recognition}, 2018, pp.
  586--595.

\bibitem{ViT}
A.~{Dosovitskiy}, L.~{Beyer}, A.~{Kolesnikov}, D.~{Weissenborn}, X.~{Zhai},
  T.~{Unterthiner}, M.~{Dehghani}, M.~{Minderer}, G.~{Heigold}, S.~{Gelly},
  J.~{Uszkoreit}, and N.~{Houlsby}, ``{An Image is Worth 16x16 Words:
  Transformers for Image Recognition at Scale},'' \emph{arXiv e-prints}, p.
  arXiv:2010.11929, Oct. 2020.

\bibitem{videogpt}
W.~{Yan}, Y.~{Zhang}, P.~{Abbeel}, and A.~{Srinivas}, ``{VideoGPT: Video
  Generation using VQ-VAE and Transformers},'' \emph{arXiv e-prints}, p.
  arXiv:2104.10157, Apr. 2021.

\end{thebibliography}

\end{document}